\documentclass{article} 
\usepackage{iclr2026_conference,times}

\usepackage{amsmath,amsfonts,bm}

\def\eqref#1{equation~\ref{#1}}

\def\1{\bm{1}}

\DeclareMathAlphabet{\mathsfit}{\encodingdefault}{\sfdefault}{m}{sl}
\SetMathAlphabet{\mathsfit}{bold}{\encodingdefault}{\sfdefault}{bx}{n}

\usepackage{hyperref}
\usepackage{url}
\usepackage{enumitem}
\usepackage{xcolor}
\usepackage{graphicx}
\usepackage{booktabs}
\usepackage{multirow}
\usepackage[table]{xcolor}
\usepackage{subcaption}
\usepackage{tcolorbox}
\usepackage{textcomp}
\usepackage[T1]{fontenc}
\tcbuselibrary{theorems,breakable,skins}
\newtcbtheorem[]{myexample}{Example}%
{
  breakable,
  enhanced,
  colback=green!5,
  colframe=green!35!black,
  fonttitle=\bfseries,
  top=2pt,         
  bottom=2pt,      
  before skip=4pt, 
  after skip=4pt   
}{th}

\title{Reasoning-Enhanced Large Language Models for Molecular Property Prediction}

\author{
    Jiaxi Zhuang$^{1,3*}$, Yaorui Shi$^{1*}$, Jue Hou$^{1*}$, Yunong He$^1$, Mingwei Ye$^1$, Mingjun Xu$^1$,\\
    \normalsize
    \textbf{Yuming Su$^1$, Linfeng Zhang$^2$, Ying Qian$^3$, Linfeng Zhang$^1$, Guolin Ke$^1$, Hengxing Cai$^{1\dagger}$} \\
    $^1$DP Technology; \quad $^2$ Shanghai Jiao Tong University; \quad $^3$ East China Normal University;\\ $^{*}$Equal contribution; \quad $^{\dagger}$Corresponding author.
}

\iclrfinalcopy 
\begin{document}

\maketitle

\begin{abstract}
Molecular property prediction is crucial for drug discovery and materials science, yet existing approaches suffer from limited interpretability, poor cross-task generalization, and lack of chemical reasoning capabilities. Traditional machine learning models struggle with task transferability, while specialized molecular language models provide little insight into their decision-making processes. To address these limitations, we propose \textbf{MPPReasoner}, a multimodal large language model that incorporates chemical reasoning for molecular property prediction.
Our approach, built upon Qwen2.5-VL-7B-Instruct, integrates molecular images with SMILES strings to enable comprehensive molecular understanding. We develop a two-stage training strategy: supervised fine-tuning (SFT) using 16,000 high-quality reasoning trajectories generated through expert knowledge and multiple teacher models, followed by Reinforcement Learning from Principle-Guided Rewards (RLPGR). RLPGR employs verifiable, rule-based rewards that systematically evaluate chemical principle application, molecular structure analysis, and logical consistency through computational verification.
Extensive experiments across 8 datasets demonstrate significant performance improvements, with MPPReasoner outperforming the best baselines by 7.91\% and 4.53\% on in-distribution and out-of-distribution tasks respectively. MPPReasoner exhibits exceptional cross-task generalization and generates chemically sound reasoning paths that provide valuable insights into molecular property analysis, substantially enhancing both interpretability and practical utility for chemists.
Code is available at \url{https://anonymous.4open.science/r/MPPReasoner-12687}.
\end{abstract}

\section{Introduction}

Molecular property prediction serves as a cornerstone in modern drug discovery and materials science, enabling researchers to computationally estimate critical molecular characteristics such as bioavailability, toxicity, and therapeutic efficacy before costly experimental validation~\citep{drugdiscoveryreview1,drugdiscoveryreview4,drugdiscoveryreview3,drugdiscoveryreview2}. Traditional experimental approaches for determining molecular properties are prohibitively expensive and time-consuming, often requiring weeks to months and costing thousands of dollars per compound~\citep{cost2,cost3,cost1}. For instance, a single ADMET (Absorption, Distribution, Metabolism, Excretion, Toxicity) screening can cost upwards of \$10,000 per molecule, making it impractical to evaluate the millions of compounds in chemical space~\citep{admetlab,admet}. This bottleneck has driven the urgent need for accurate and efficient computational models that can predict molecular properties at scale~\citep{recenttrends}

Despite decades of research, current molecular property prediction approaches face fundamental limitations that hinder their practical adoption. Early computational methods relied on hand-crafted molecular descriptors and traditional machine learning algorithms, which struggle with task transferability and require extensive feature engineering for each new application~\citep{quantumdescriptors,extendedfingerprints,moleculenet}. More recent advances have introduced specialized molecular models, including graph neural networks (GNNs)~\citep{crystalgnn,unimol} and molecular language models~\citep{chemberta,biot5-plus,moleculargpt}, which have achieved impressive performance by learning molecular representations directly from graph structures or SMILES strings. However, these approaches suffer from a critical limitation: they lack interpretability and fail to provide chemically meaningful explanations for their predictions. When a model predicts that a molecule is toxic, chemists cannot understand which structural features or chemical principles led to this conclusion, limiting the model's utility in real-world drug development pipelines where understanding the rationale behind predictions is crucial for decision-making.

The fundamental limitation shared by all existing molecular property prediction methods is the absence of \textbf{effective chemical reasoning}—the ability to analyze molecular structures, identify relevant functional groups, apply chemical principles, and provide coherent explanations for property predictions. When experienced chemists evaluate molecular properties, they follow a structured reasoning process: examine the molecular structure to identify key functional groups (e.g., hydroxyl groups affecting solubility~\citep{organic}), consider relevant chemical principles (e.g., Lipinski's Rule for drug-likeness~\citep{fiverule}), analyze structure-activity relationships~\citep{sartoxic}, and synthesize these insights to make informed predictions~\citep{augmenting,xai}. This reasoning capability is crucial not only for accuracy but also for trust and adoption in practical applications, as chemists need to understand the rationale behind predictions to make informed decisions about lead compound optimization and safety assessments.

To address these fundamental limitations, we propose MPPReasoner, a novel multimodal framework that successfully introduces chemical reasoning capabilities into molecular property prediction. MPPReasoner represents the first systematic attempt to cultivate domain-specific reasoning abilities for molecular property prediction, enabling models to analyze molecular structures, apply established chemical principles, and provide human-interpretable explanations during the prediction process. Built upon Qwen2.5-VL-7B-Instruct~\citep{qwen25vl}, MPPReasoner integrates multimodal molecular representations by combining 2D molecular images with SMILES strings, enabling comprehensive structural understanding from both visual and textual modalities. 
Our training methodology employs a two-stage strategy to progressively develop chemical reasoning capabilities: 1) Supervised Fine-Tuning (SFT) with carefully curated reasoning trajectories generated through expert knowledge and teacher models, establishing foundational reasoning patterns; 2) Reinforcement Learning from Principle-Guided Rewards (RLPGR), a novel reward framework that leverages verifiable, rule-based feedback to enhance chemical reasoning quality. Unlike traditional reinforcement learning (RL) approaches, RLPGR decomposes chemical reasoning into hierarchical reward components that evaluate logical consistency, chemical principle application accuracy, and molecular structure analysis precision through computational verification.

Extensive experiments on 8 diverse molecular property prediction datasets demonstrate the effectiveness of our approach, achieving substantial performance improvements with average ROC-AUC scores~\citep{roc,auc} of 0.8068 on in-distribution (ID) tasks and 0.7801 on out-of-distribution (OOD) tasks, outperforming the best existing baselines by 7.91 and 4.53 percentage points respectively. Notably, our model exhibits exceptional OOD generalization capabilities, with particularly significant improvements on OOD datasets where many specialist models lack evaluation capability. Through expert evaluation and detailed case studies, we demonstrate that our approach produces chemically sound explanations that provide valuable insights into molecular property relationships.
The main contributions of this work are as follows:

\begin{itemize}[leftmargin=*]
    \item We successfully introduce chemical reasoning capabilities into molecular property prediction tasks through MPPReasoner, representing a systematic approach to enable structured analysis, chemical principles application, and mechanistic explanations during the prediction process.
    
    \item We propose a comprehensive training strategy that combines high-quality reasoning trajectories SFT and RLPGR, a novel hierarchical reward framework targeting chemical reasoning quality through verifiable, rule-based feedback on logical consistency, comparative analysis, chemical principle usage and molecular structure analysis .
    
    \item We construct a carefully curated dataset of chemical reasoning trajectories generated through expert knowledge and few-shot prompting with multiple teacher models, providing a valuable resource for training reasoning-capable molecular property prediction models.
    
    \item We demonstrate significant performance improvements across 8 datasets with superior OOD generalization, while providing enhanced interpretability through expert-validated reasoning paths that offer insights for chemists in real-world applications.
\end{itemize}

\section{Related Work}
\label{sec:background}
This section reviews prior research on machine learning for molecular representations, multimodal language models in chemistry, and reasoning capabilities in LLMs, which are foundational to our proposed framework for training reasoning LLMs tailored to molecular property prediction.

\paragraph{Machine Learning for Molecular Representation.}
GNNs have evolved as a dominant paradigm for molecular graph representation, progressing from early convolutional \citep{moleculargraph,schnet} and message-passing \citep{mpnn} frameworks to sophisticated 3D-aware models \citep{mpnn, rgcl, simsgt,unimol}, enabling robust applications in property prediction, virtual screening, and drug discovery \citep{gnn-chemistry-applications}.
In parallel, specialized molecular language models have reframed molecular structures as textual sequences such as SMILES strings\citep{smiles}, with models like MolecularGPT \citep{moleculargpt} and BioT5-Plus \citep{biot5-plus} supporting few-shot adaptation and multi-task learning for diverse chemical and biological tasks \citep{scilitllm, reactxt, molca}. 

\paragraph{Multimodal Language Models for Chemistry.}
The emerging trend of multimodal LLMs in chemistry integrates diverse data types—such as SMILES strings and molecular graphs to address unimodal limitations, as seen in foundational molecular-text models \citep{mol-llm, chemvlm, molca}, instruction-tuned assistants \citep{instructmol}, and tool-augmented systems \citep{chemcrow}, enhancing robustness in property prediction \citep{molt5}, molecular design \citep{mol-llm}, and synthesis planning \citep{relm, reactxt}. However, these models still lack the capability to provide chemical reasoning for their predictions.

\paragraph{Reasoning in Large Language Models.}
Reasoning capabilities have demonstrated remarkable efficacy in commercial LLMs, particularly through chain-of-thought processes as exemplified in OpenAI's o1 series and other advanced models \citep{cot,openai-o1,gemini,claude}.
Training these abilities leverages RL techniques, from Proximal Policy Optimization \citep{ppo} in RL from Human Feedback (RLHF) \citep{rlhf} for preference alignment, to efficient extensions like Group Relative Policy Optimization (GRPO) \citep{grpo} with outcome-based rewards and Reinforcement Learning with Verifiable Rewards (RLVR) for one-shot verifiable steps, improving generalization on complex tasks \citep{rlvr}.
These RL advancements motivate our adaptation for chemical-specific reasoning in the field of molecular property prediction.
\section{Methodology}
\label{sec3:method}

\begin{figure}
  \centering
  \includegraphics[clip, trim=0cm 0.5cm 0cm 0.5cm, width=0.97\linewidth]{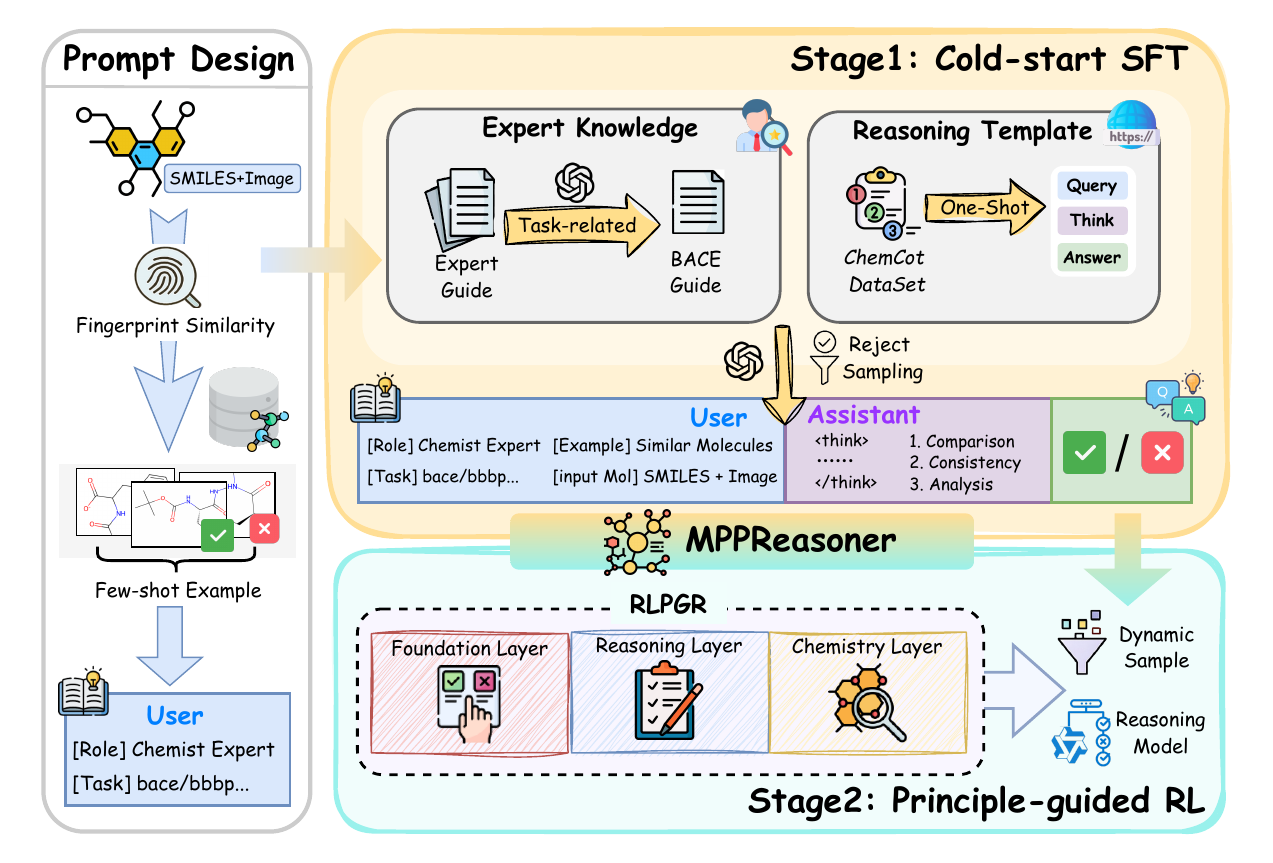}
  \caption{ Overview of MPPReasoner framework.} 
  \label{fig:framework}
\end{figure}

MPPReasoner cultivates chemical reasoning capabilities in multimodal large language models through a structured approach illustrated in Figure~\ref{fig:framework}. We first construct high-quality reasoning trajectories that demonstrate expert-level chemical analysis patterns. These trajectories are then used in a two-stage training framework: SFT establishes foundational reasoning abilities, followed by RL guided by our novel Principle-Guided Reward mechanism.

\subsection{Multimodal Molecular Prompt Design}

To provide comprehensive molecular understanding, we employ a multimodal input representation that integrates 2D molecular images with their corresponding SMILES strings. This dual representation enables the model to capture both the sequential chemical information encoded in SMILES and the spatial structural relationships depicted in molecular visualizations.

As shown in Appendix~\ref{appa:prompt}, our prompt engineering strategy comprises four essential components: \textbf{[Role Definition]} instructs the model to act as an expert chemist specializing in molecular property prediction; \textbf{[Task Description]} provides task-specific instructions outlining the prediction objective and requirement for step-by-step reasoning; \textbf{[Few-Shot Examples]} are dynamically retrieved by identifying the top-5 most similar molecules from the training set using Tanimoto similarity~\citep{tanimoto} based on Morgan fingerprints~\citep{fingerprint}; and \textbf{[Multimodal Molecule]} includes both the rendered 2D molecular structure image and the corresponding SMILES string, providing complementary perspectives on molecular characteristics.

\subsection{Two-Stage Training Framework for Chemical Reasoning}

\subsubsection{Stage 1: Reasoning Trajectory Construction for Cold Start SFT}
\label{sec3.2.1:stage1}
The first stage establishes foundational chemical reasoning capabilities through high-quality data construction and supervised learning. We begin by constructing a comprehensive dataset of chemical reasoning trajectories through two complementary approaches:

\paragraph{Multi-Source Reasoning Data Construction.} We leverage powerful large general-domain reasoning models as teacher models to generate high-quality chemical reasoning patterns through two complementary approaches:
\begin{itemize}[leftmargin=*]
    \item \textit{ChemCoT-Based One-Shot Generation:} We utilize exemplars from the ChemCoT dataset~\citep{chemcot} as one-shot demonstrations, instructing teacher models to emulate the step-by-step analytical style demonstrated in chemical chain-of-thought examples.
    \item \textit{Expert-Guided Task-Specific Generation:} Expert chemists draft comprehensive reasoning guides covering fundamental principles for various molecular properties. These guides are then refined by GPT-4o~\citep{gpt4o} to extract task-specific knowledge relevant to each dataset (e.g., BACE~\citep{moleculenet}). The extracted principles serve as domain-specific prompts in generating reasoning trajectories that incorporate relevant chemical knowledge and theoretical foundations.
\end{itemize}

\paragraph{Quality Control and Data Curation.} We employ rejection sampling to ensure trajectory quality, accepting only those instances where teacher models produce correct predictions (True/False). This process yields 16,000 high-quality reasoning trajectories for SFT.

Using these curated reasoning trajectories, we perform SFT on Qwen2.5-VL-7B-Instruct with standard next-token prediction loss. The reasoning generation process utilizes three state-of-the-art language models (GPT4o~\citep{gpt4o}, DeepSeek-v3.1~\citep{deepseekr1} and Qwen2.5VL~\citep{qwen25vl}) working in parallel to ensure diversity in reasoning patterns while maintaining high quality through model complementarity. When multiple teacher models generate correct reasoning for the same instance, we randomly select one trajectory to maintain dataset diversity. This stage focuses on instruction alignment, teaching the model to follow the required format of providing step-by-step reasoning before making final predictions, while simultaneously instilling domain-specific knowledge and analytical patterns demonstrated in the expert-curated trajectories.

\subsubsection{Stage 2: Advanced Reasoning Refinement with RLPGR}
\begin{figure}
  \centering
  \includegraphics[width=\linewidth]{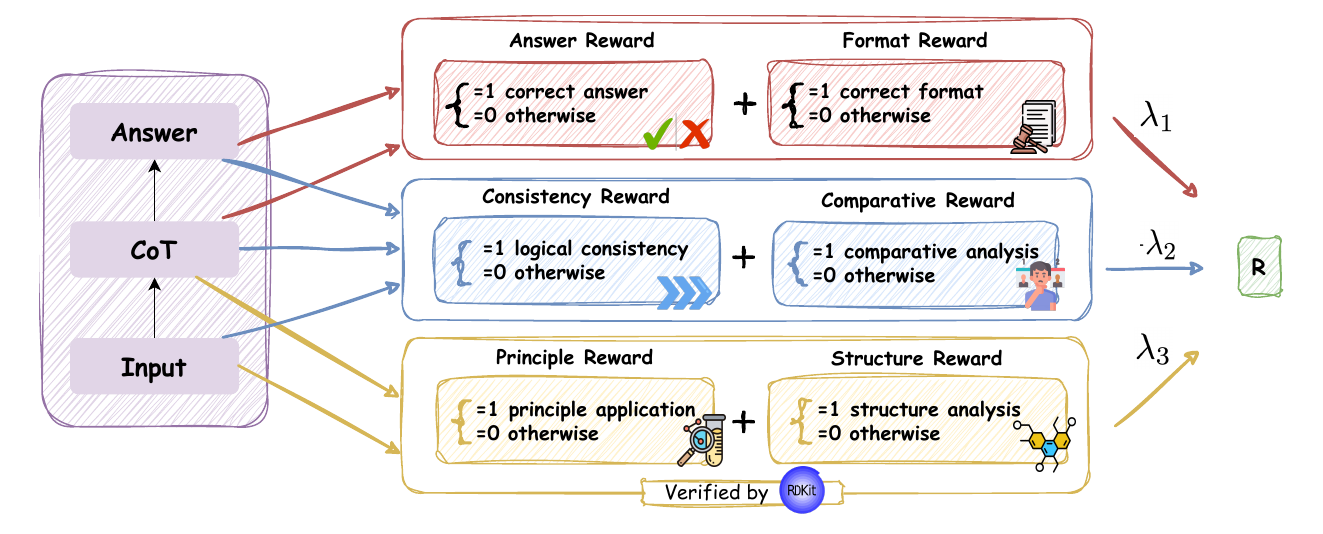}
  \caption{Illustration of RLPGR in MPPReasoner.} 
  \label{fig:rlpgr}
\end{figure}

While SFT establishes basic reasoning patterns, the second stage employs RLPGR to elevate the model's capabilities from imitation to exploration and refinement. Unlike traditional RLHF~\citep{rlhf} approaches that rely on human preference data, RLPGR leverages verifiable, rule-based rewards~\citep{rlvr} derived from chemical principles and computational tools~\citep{rdkit}, ensuring both scalability and domain accuracy.

Our RLPGR framework decomposes the complex cognitive process of chemical reasoning into measurable reward components across three hierarchical layers. As illustrated in Figure~\ref{fig:rlpgr}, given a molecular description $x$, reasoning trace $z$, and prediction $y$, the total reward is computed as:


\begin{equation}
R_{\text{total}}(x,z,y) = \lambda_1 \underbrace{(r_{\text{ans}} + r_{\text{fmt}} 
)}_{R_{\text{foundation}}} + \lambda_2 \underbrace{(r_{\text{cons}} + r_{\text{comp}})}_{R_{\text{reasoning}}} + \lambda_3 \underbrace{(r_{\text{prin}} + r_{\text{struct}})}_{R_{\text{chemistry}}}
\end{equation}

where $\lambda_i$ are hyperparameters controlling the relative importance of each reward component.

\textbf{Foundation Layer:} This layer ensures basic task requirements through two components:
\begin{itemize}[leftmargin=*]
  \item Answer reward $r_{\text{ans}}$ provides binary feedback based on prediction correctness.
  \item Format compliance reward $r_{\text{fmt}}$ verifies that outputs follow the required structure with reasoning enclosed in \texttt{<think>} tags and predictions in \texttt{<answer>} tags.
\end{itemize}

\textbf{Reasoning Layer:} This layer evaluates general reasoning quality through two key aspects:
\begin{itemize}[leftmargin=*]
  \item Logical consistency reward $r_{\text{cons}}$ measures alignment between the reasoning conclusion and final prediction by analyzing sentiment consistency using predefined keyword sets for affirmative and negative conclusions.
  \item Comparative analysis reward $r_{\text{comp}}$ encourages analogical thinking by detecting whether the reasoning process analyzes the few-shot examples provided based on molecular similarity, promoting effective utilization of the retrieved similar molecules and cross-molecular reasoning capabilities.
\end{itemize}

\textbf{Chemistry Layer:} This layer targets domain-specific expertise through computational verification of chemical knowledge and structural analysis accuracy. We leverage RDKit~\citep{rdkit} for molecular property computation and substructure detection to provide objective feedback on chemical reasoning quality.
\begin{itemize}[leftmargin=*]
  \item Chemical principle application reward $r_{\text{prin}}$ evaluates whether mentioned chemical concepts align with computationally derived molecular properties. For instance, when the reasoning discusses hydrophobicity, we verify this against the computed LogP value. This reward ensures that chemical principles are applied appropriately rather than superficially mentioned.
  \item Molecular structure analysis reward $r_{\text{struct}}$ measures the coverage of structural feature identification as 
  $
  r_{\text{struct}} = {|S_{\text{actual}} \cap S_{\text{pred}}|}/({|S_{\text{actual}}| + \epsilon}),
  $
  where $S_{\text{actual}}$ represents the set of distinct structural feature types identified via RDKit (functional groups, ring systems, stereochemical features), $S_{\text{pred}}$ denotes the set of structural feature types mentioned in the reasoning trace $z$, and $\epsilon = 10^{-5}$ ensures numerical stability.
\end{itemize}

\textbf{Training Process.} We employ GRPO \citep{grpo} for policy optimization, which maximizes the expected reward across reasoning trajectories:

\begin{equation}
\mathcal{L}_{\text{RLPGR}} = \mathbb{E}_{(x,z,y) \sim \pi_\theta} [R_{\text{total}}(x,z,y)]
\end{equation}

where $\pi_\theta$ represents the policy parameterized by $\theta$. Dynamic sampling during training focuses computational resources on tractable reasoning examples, ensuring efficient convergence toward models capable of generating chemically accurate and interpretable reasoning paths. This RLPGR approach transforms the model from pattern-matching to genuine chemical reasoning through systematic principle-guided RL.

\section{Experiments}

To comprehensively evaluate our approach, we investigate the following Research Questions (RQs):

\textbf{RQ1:} Does MPPReasoner achieve superior performance compared to existing molecular property prediction methods on both ID and OOD datasets?

\textbf{RQ2:} What are the individual contributions of the two-stage training strategy and the RLPGR reward components to the overall performance and reasoning quality?

\textbf{RQ3:} Can our model generate high-quality reasoning paths that provide chemically meaningful insights comparable to expert-level analysis?

\subsection{Experimental Setup}
\label{sec4.1:setup}

\paragraph{Datasets.} 
We evaluate MPPReasoner on 8 diverse molecular property prediction datasets to assess both ID and OOD performance. The datasets are categorized as follows:
\begin{itemize}[leftmargin=*]
\item \textit{ID Datasets:} We utilize four benchmark datasets from MoleculeNet~\citep{moleculenet}, which is widely used to predict whether the given molecule has specific properties: BACE (1,513), BBBP (2,039), SIDER (1,427), HIV (41,127).

\item \textit{OOD Datasets:} We employ four datasets from the Therapeutic Data Commons (TDC)~\citep{tdc1,tdc2} to evaluate cross-task generalization capabilities: Bioavailability (128), CYP2C9\_V (2,418), CYP2D6\_V (2,626), AMES (1,456).
\end{itemize}

The ID/OOD categorization is based on whether the training set includes samples from the corresponding dataset. For training, we randomly sample 4,000 instances from the ID datasets to ensure balanced representation across different molecular properties. Test sets follow standard benchmarking protocols established in prior literature to maintain fair comparison with baseline methods.

\paragraph{Baseline.}
We compare MPPReasoner against two categories of approaches:
\begin{itemize}[leftmargin=*]
\item \textit{Task-specific Specialist Models:} These models are designed for molecular property prediction: Graphormer-p~\citep{graphformer}, Uni-Mol~\citep{unimol}, GIMLET~\citep{gimlet}, MolecularGPT~\citep{moleculargpt}, Mol-LLM~\citep{mol-llm}, InstructMol-GS~\citep{instructmol}, BioT5-Plus~\citep{biot5-plus}, MolXPT~\citep{molxpt}.

\item \textit{LLM-based Generalist Models:} These include reasoning models: o3-mini~\citep{o3mini}, DeepSeek-V3.1~\citep{deepseekr1}, large-scale models: GPT-4o~\citep{gpt4o}, Qwen2.5-VL-72B-Instruct~\citep{qwen25vl}, and baseline models: Qwen2.5-VL-7B-Instruct~\citep{qwen25vl} applied to molecular property prediction.
\end{itemize}

Implementation details and hyper-parameters settings are provided in Appendix~\ref{appc:setting}.

\begin{table}
\setlength{\tabcolsep}{4pt}
\caption{Performance comparison of task-specific specialist models and LLM-based generalist models on ID and OOD benchmarks. Best performance is in \textbf{bold}. }
\label{tab:mol_results}
\centering
\resizebox{\textwidth}{!}{%
\begin{tabular}{lcccc|cccc|cc}
\toprule
\multicolumn{1}{c}{\multirow{2.5}{*}{\textbf{Model}}} 
& \multicolumn{4}{c|}{\textbf{ID Performance}} 
& \multicolumn{4}{c|}{\textbf{OOD Performance}} 
& \multicolumn{2}{c}{\textbf{Average}} \\
\cmidrule(lr){2-5} \cmidrule(lr){6-9} \cmidrule(lr){10-11}
& BACE & BBBP & SIDER & HIV 
& Bioavail. & C2C9\_V & C2D6\_V & AMES
& ID & OOD \\
\midrule
\rowcolor{gray!20}
\multicolumn{11}{c}{\textit{\# Task-specific specialist models}} \\
Graphormer-p   & 0.8575 & 0.7163 & --     & 0.7788 
               & -- & -- & -- & --   
               & 0.7842 & -- \\
Uni-Mol        & 0.8570 & 0.7290 & 0.6590 & \textbf{0.8080} 
               & -- & -- & -- & --   
               & 0.7633 & -- \\
GIMLET         & 0.6957 & 0.5939 & --     & 0.6624 
               & -- & -- & -- & --   
               & 0.6507 & -- \\
MolecularGPT   & 0.7331 & 0.6822 & --     & 0.6382 
               & -- & -- & -- & --   
               & 0.6845 & -- \\
Mol-LLM        & 0.8080 & \textbf{0.8430} & \underline{0.7610} & 0.7650 
               & -- & -- & -- & --   
               & 0.7943 & -- \\
InstructMol-GS & 0.8210 & 0.7240 & --     & 0.6890 
               & -- & -- & -- & --   
               & 0.7447 & -- \\
BioT5-Plus     & 0.8620 & 0.7650 & 0.5201 & 0.7630 
               & 0.5243 & 0.4971 & 0.5321 & 0.4466 
               & 0.7275 & 0.5000 \\
MolXPT         & \underline{0.8840} & \underline{0.8000} & 0.7170 & 0.7810 
               & 0.4749 & 0.5904 & 0.5291 & 0.6073 
               & \underline{0.7955} & 0.5504 \\
\midrule
\rowcolor{gray!20}
\multicolumn{11}{c}{\textit{\# LLM-based generalist models}} \\
o3-mini        & 0.7891 & 0.5972 & 0.5626 & 0.6039 
               & 0.6246 & \underline{0.7729} & \underline{0.7643} & \underline{0.8361} 
               & 0.6382 & \underline{0.7495} \\
DeepSeek-V3.1-Think  & 0.7017 & 0.6048 & 0.5637 & 0.5938 
               & \underline{0.6572} & 0.7633 & 0.7484 & 0.8218 
               & 0.6160 & 0.7477 \\
GPT-4o         & 0.6070 & 0.6731 & 0.6347 & 0.5698 
               & 0.5826 & 0.5508 & 0.5902 & 0.6141 
               & 0.6212 & 0.5844 \\
Qwen2.5-VL-72B-Instruct & 0.7764 & 0.5791 & 0.5880 & 0.7325 
               & 0.6388 & 0.7624 & 0.7222 & 0.8156 
               & 0.6690 & 0.7348 \\ 
Qwen2.5-VL-7B-Instruct  & 0.6910 & 0.6175 & 0.5823 & 0.5125 
               & 0.5232 & 0.7333 & 0.6999 & 0.7667 
               & 0.6008 & 0.6808 \\
\rowcolor{gray!15}
\textbf{MPPReasoner (Ours)} & \textbf{0.9090} & 0.7436 & \textbf{0.8280} & \underline{0.7932} & \textbf{0.6728} & \textbf{0.8480} & \textbf{0.7950} & \textbf{0.8750} & \textbf{0.8190} & \textbf{0.7977} \\
\bottomrule
\end{tabular}}
\end{table}

\subsection{Main Results (RQ1)}

Table~\ref{tab:mol_results} presents the comprehensive performance comparison of MPPReasoner against state-of-the-art baselines across all 8 datasets.
On ID datasets, MPPReasoner demonstrates competitive performance with specialized models while maintaining the advantage of using a smaller 7B parameter architecture. MPPReasoner achieves the best performance on challenging tasks like BACE and SIDER, indicating successful capture of complex molecular property relationships such as enzyme inhibition and side effect prediction. While some specialized models like Mol-LLM excel on specific tasks such as BBBP, these models achieve high ID performance at the expense of cross-task adaptability, completely lacking OOD evaluation capability. This specialization-generalization trade-off limits their practical utility in real-world scenarios requiring diverse molecular property assessment.

The most significant advantage emerges in OOD scenarios, where MPPReasoner substantially outperforms all baseline categories by 6.43\% over the best reasoning model. This consistent superiority across diverse molecular property types highlights how domain-specific chemical reasoning outperforms both general reasoning capabilities and raw parameter scaling approaches. The performance gap becomes even more pronounced when considering that MPPReasoner operates with significantly fewer parameters than competing large-scale models.

The results reveal fundamental differences between model categories and their limitations. Task-specific specialist models excel in familiar scenarios but completely lack cross-task generalization capabilities, while generalist models show consistent cross-task performance but suffer from insufficient domain expertise. MPPReasoner uniquely bridges this gap by embedding domain-specific reasoning rather than relying on general reasoning patterns or parameter scaling alone. The transformative impact becomes evident when comparing MPPReasoner to its base model, showing dramatic improvements of 36.36\% on ID tasks and 17.17\% on OOD tasks. This demonstrates that structured chemical reasoning fundamentally enhances molecular understanding beyond conventional approaches, enabling both specialist-level accuracy and generalist-level adaptability through systematic integration of chemical principles.

\subsection{Ablation Studies (RQ2)}

\begin{table}
\setlength{\tabcolsep}{4pt}
\caption{Ablation study on training stages and RLPGR reward. }
\label{tab:ablation_study}
\centering
\resizebox{\textwidth}{!}{%
\begin{tabular}{lcccc|cccc|cc}
\toprule
\multicolumn{1}{c}{\multirow{2.5}{*}{\textbf{Setting}}} & \multicolumn{4}{c|}{\textbf{ID Performance}} & \multicolumn{4}{c|}{\textbf{OOD Performance}} & \multicolumn{2}{c}{\textbf{Average}} \\
\cmidrule(lr){2-5} \cmidrule(lr){6-9} \cmidrule(lr){10-11}
& BACE & BBBP & SIDER & HIV 
& Bioavail. & C2C9\_V & C2D6\_V & AMES
& ID & OOD \\
\midrule
Base (Qwen2.5-VL-7b-Instruct) & 0.6910 & 0.6175 & 0.5823 & 0.5125 & 0.5232 & 0.7333 & 0.6999 & 0.7667 & 0.6008 & 0.6808 \\
\midrule
SFT Only & 0.8558 & 0.6824 & 0.6752 & 0.7186 & 0.6625 & 0.7799 & 0.7348 & 0.8415 & 0.7330 & 0.7547 \\
RL Only (RLPGR) & 0.8142 & 0.5733 & 0.7428 & 0.5552 & 0.6632 & 0.7491 & 0.6732 & 0.7300 & 0.6714 & 0.7039 \\
\midrule
SFT + $R_{\text{foundation}}$ & 0.8836 & 0.6794 & 0.8089 & 0.7556 & 0.6358 & 0.8364 & 0.7862 & 0.8536 & 0.7819 & 0.7780 \\
SFT + $R_{\text{foundation}}$ + $R_{\text{reasoning}}$ & 0.8877 & 0.7104 & 0.7981 & 0.7560 & \textbf{0.6771} & 0.8140 & 0.7795 & 0.8388 & 0.7881 & 0.7774 \\
\rowcolor{gray!15}
\textbf{MPPReasoner (Ours)} & \textbf{0.9090} & \textbf{0.7459} & \textbf{0.8280} & \textbf{0.7932} & 0.6728 & \textbf{0.8480} & \textbf{0.7950} & \textbf{0.8750} & \textbf{0.8190} & \textbf{0.7977} \\
\bottomrule
\end{tabular}}
\end{table}

To understand the individual contributions of our two-stage training strategy and the hierarchical reward components in RLPGR, we conduct comprehensive ablation studies. Table~\ref{tab:ablation_study} presents the systematic analysis of each component's impact on both ID and OOD performance.
The results demonstrate that both SFT and RL stages contribute significantly to overall performance, with distinct advantages for different aspects. SFT alone provides substantial improvements over the base model, achieving 22.01\% and 10.85\% gains on ID and OOD tasks respectively, indicating that SFT with high-quality reasoning trajectories successfully instills foundational chemical reasoning capabilities. RL alone also shows meaningful improvements of 11.74\% and 3.39\%, demonstrating that principle-guided rewards can enhance reasoning quality independently. However, the combination of SFT + RL yields the strongest performance with 36.36\% and 17.17\% improvements, revealing important synergistic effects between two training stages that exceed their individual contributions.

The progressive addition of RLPGR reward components shows clear incremental benefits, validating our hierarchical design. Foundation rewards provide substantial improvements of 6.67\% on ID and 3.09\% on OOD over SFT alone, establishing enhanced task completion capabilities beyond basic instruction following. The Reasoning layer contributes additional 0.84\% ID gains while maintaining similar OOD performance, indicating that logical consistency and comparative analysis refine reasoning quality without compromising generalization. Most importantly, the Chemistry layer delivers the largest incremental improvements of 3.92\% on ID and 2.61\% on OOD, confirming that domain-specific chemical principle verification is crucial for molecular property prediction tasks.

\subsection{Reasoning Quality Evaluation (RQ3)}
\label{sec4.4:quality}
\begin{figure}
    \centering
    \begin{subfigure}[b]{0.49\textwidth}
        \centering
        \includegraphics[width=\textwidth]{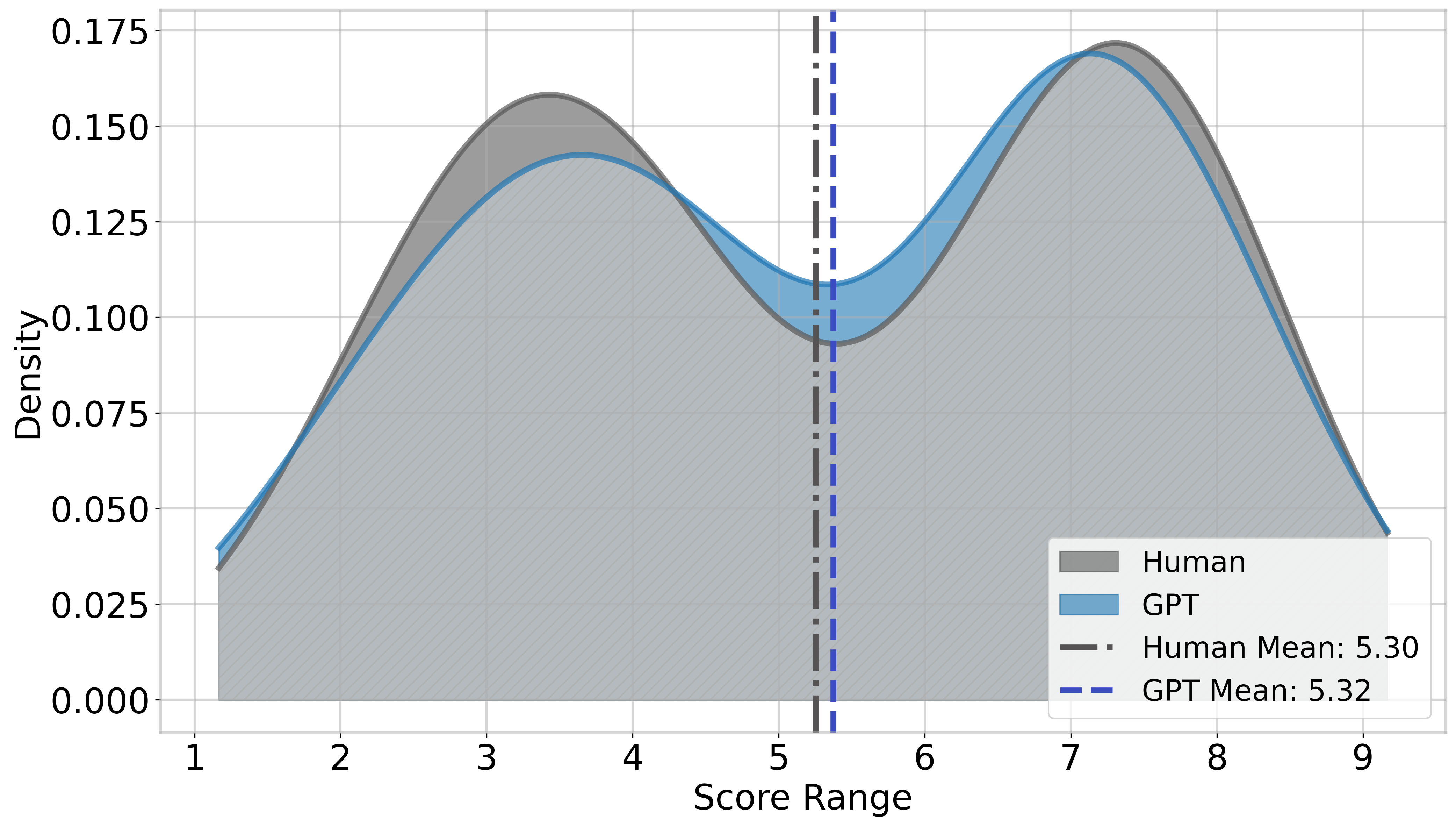}
        \caption{Human-AI evaluation consistency}
        \label{fig:score_distribution}
    \end{subfigure}
    \hfill
    \begin{subfigure}[b]{0.49\textwidth}
        \centering
        \includegraphics[width=\textwidth]{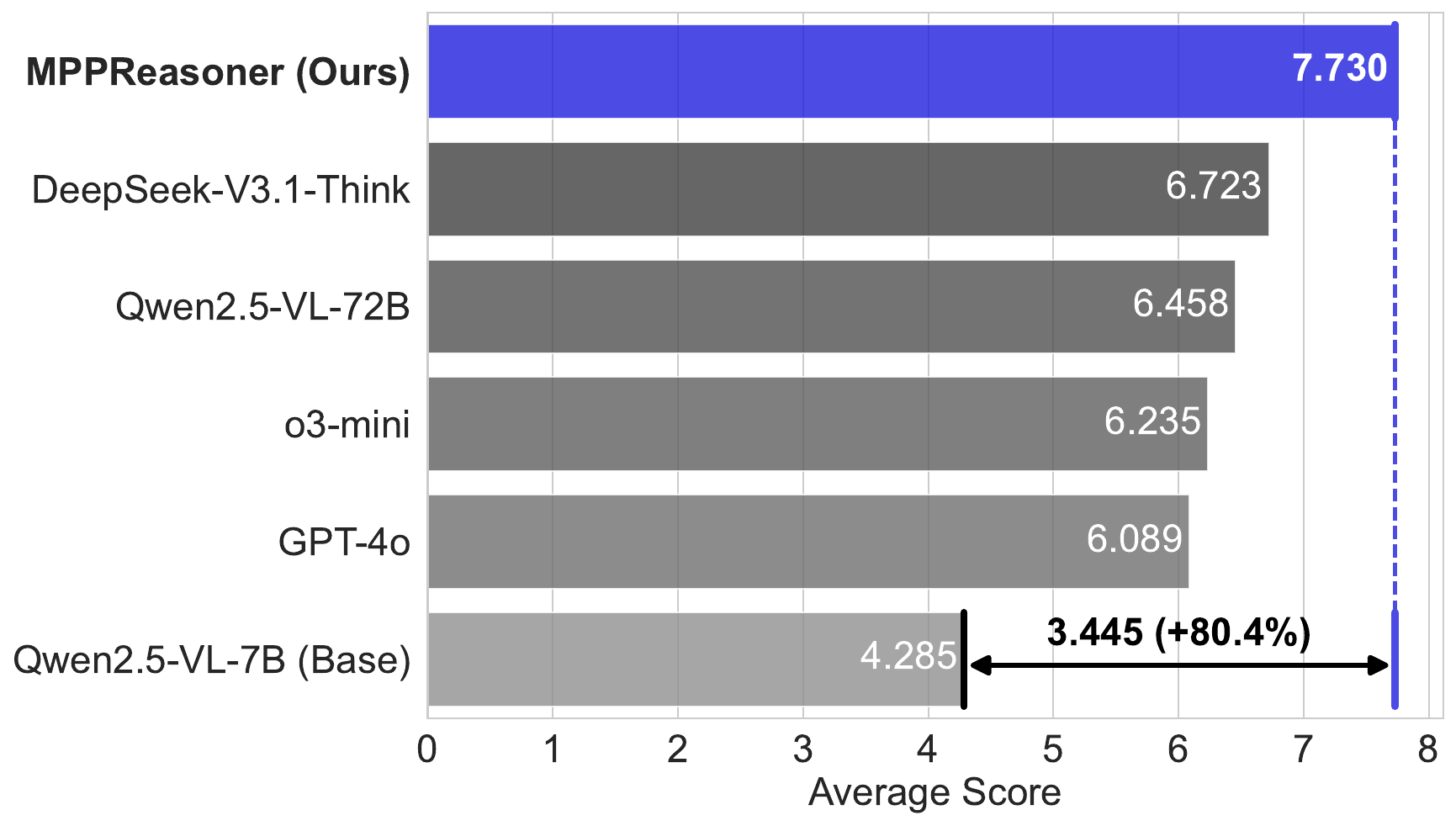}
        \caption{Model reasoning quality scores}
        \label{fig:model_performance}
    \end{subfigure}
    \vspace{-1em}
    \caption{Reasoning quality evaluation results. (a) Strong consistency between automated and human assessments with $\rho$ = 0.82. (b) MPPReasoner achieves the highest reasoning quality score.}
    \label{fig:reasoning_evaluation}
\end{figure}

To assess whether MPPReasoner generates high-quality reasoning paths that provide chemically meaningful insights, we conduct systematic evaluation using automated assessment validated against human expert judgment. We employ a LLM-as-a-Judge~\citep{llmasjudge} framework using GPT-4o to evaluate three dimensions~\citep{sophiavlr1}: \textbf{logical soundness, accuracy \& insight, and conciseness}, each scored on a 0-10 scale with detailed rubrics. To establish reliability, we validate GPT-4o scores against human expert assessments on 60 reasoning samples from three baseline models. Figure~\ref{fig:reasoning_evaluation}(a) shows remarkable consistency between automated and human evaluations, with similar distributions and central tendencies with spearman correlation coefficient reaches $\rho$ = 0.82.

Figure~\ref{fig:reasoning_evaluation}(b) presents the comparative reasoning quality assessment across different model categories, showing average scores across the three evaluation dimensions. MPPReasoner achieves the highest score of 7.730, substantially outperforming advanced reasoning models including DeepSeek-V3.1-Think at 6.723 and o3-mini at 6.235, as well as large-scale models like Qwen2.5-VL-72B at 6.458 and GPT-4o at 6.089. Despite using a smaller 7B architecture, MPPReasoner demonstrates 15.0\% improvement over the best baseline, highlighting how domain-specific chemical reasoning surpasses both general reasoning capabilities and parameter scaling approaches. The superior reasoning quality translates to practical benefits: MPPReasoner consistently identifies relevant functional groups, applies appropriate chemical principles, and provides mechanistic explanations that enable chemists to understand both what properties a molecule has and why these properties emerge from specific structural features. Detailed dimensional scores are provided in Appendix~\ref{appb:quality}.

\section{Case Study}


\begin{figure}
    \centering
    \includegraphics[width=0.98\textwidth]{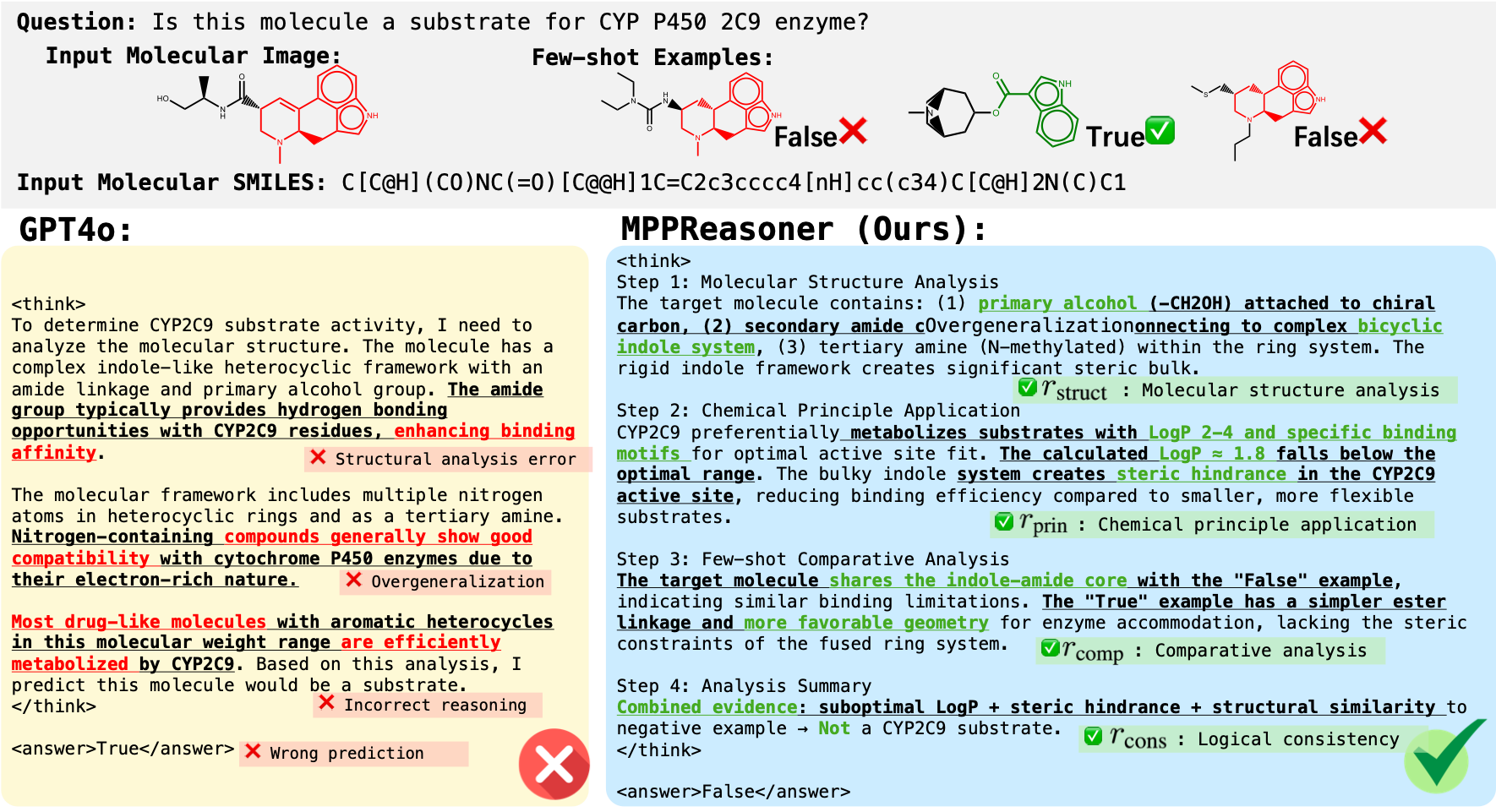}
    \caption{Case study comparison between GPT-4o and MPPReasoner for CY-P450-2C9 substrate prediction. The figure demonstrates how RLPGR training enables systematic chemical reasoning with accurate predictions, while GPT4o suffer from several errors.}
    \label{fig:case_study}
\end{figure}

To illustrate the practical benefits of our chemical reasoning approach, Figure~\ref{fig:case_study} presents a representative case study comparing MPPReasoner with GPT-4o on CY-P450-2C9 substrate prediction. The comparison reveals fundamental differences in analytical quality: GPT-4o exhibits critical reasoning flaws including structural analysis errors (incorrectly assuming amide groups enhance binding affinity), overgeneralization (broadly claiming nitrogen-containing compounds show P450 compatibility), and incorrect reasoning patterns (unsupported statistical generalizations), ultimately leading to a wrong prediction. In contrast, MPPReasoner demonstrates systematic chemical reasoning through accurate molecular structure analysis (precise functional group identification), correct chemical principle application (referencing CYP2C9-specific LogP requirements and calculating steric hindrance), meaningful comparative analysis (connecting structural similarities to substrate labels), and logical consistency (integrating multiple evidence sources). This exemplifies how RLPGR's hierarchical rewards successfully cultivate domain-specific reasoning capabilities, enabling chemists to trust the model's mechanistic insights for practical applications.

\section{Conclusion}

This work introduces MPPReasoner, a novel multimodal large language model that systematically incorporates chemical reasoning for molecular property prediction. Through a novel two-stage training strategy combining supervised fine-tuning with RLPGR, our approach successfully bridges the gap between specialist accuracy and generalist adaptability. Extensive experiments across 8 datasets demonstrate substantial performance improvements of 20.60\% and 9.93\% on ID and OOD tasks respectively, with exceptional cross-task generalization capabilities where many specialist models lack evaluation capability. Beyond performance gains, MPPReasoner generates chemically sound reasoning paths that enable chemists to understand not just prediction outcomes but the underlying chemical rationale. This represents a crucial advancement toward interpretable AI systems that provide mechanistic insights grounded in established chemical principles, supporting informed decision-making in drug discovery and offering a potential blueprint for other scientific domains.
\section*{Ethics Statement}

In developing MPPReasoner, we prioritized ethical considerations to ensure responsible use of our models and methodologies. Our research does not involve human subjects, and all molecular data are obtained from publicly available, copyright-compliant datasets (MoleculeNet and TDC) with appropriate research licensing. We employed rigorous quality filtering processes during data curation to minimize biased or misleading chemical information.
We acknowledge that biases inherent in molecular property datasets, including overrepresentation of certain chemical scaffolds, may propagate into model outputs. While MPPReasoner is designed for beneficial applications in drug discovery and materials science, we encourage responsible use within established scientific and regulatory frameworks. Our work is conducted with commitment to research integrity, ensuring contributions remain beneficial to the scientific community while addressing ethical responsibilities of developing AI technologies for chemical applications.

\section*{Reproducibility Statement}
We have made comprehensive efforts to ensure the reproducibility of MPPReasoner and our experimental findings. Our two-stage training methodology is detailed in Section~\ref{sec3:method}, including the SFT process and the novel RLPGR framework with specific reward components. Complete implementation details, hyperparameter configurations, and training procedures are provided in Appendix~\ref{appc:setting}. The experimental setup, including dataset descriptions, baseline model configurations, and evaluation protocols, is thoroughly documented in Section~\ref{sec4.1:setup} and Appendix~\ref{appc:setting}.
All datasets used in our experiments are publicly available: the ID datasets are from MoleculeNet, and the OOD datasets are from TDC. The reasoning trajectory construction process using expert knowledge and multiple teacher models is described in Section~\ref{sec3.2.1:stage1}, with specific prompting strategies detailed in Appendix~\ref{appa:prompt}. Our reasoning quality evaluation methodology, including the LLM-as-a-Judge framework and human expert validation procedures, is documented in Section~\ref{sec4.4:quality}. We provide the complete source code for model training, evaluation, and reasoning quality assessment at \url{https://anonymous.4open.science/r/MPPReasoner-12687}.

\bibliography{arxiv}
\bibliographystyle{iclr2026_conference}

\newpage
\appendix

\section{Prompt Templates}
\label{appa:prompt}
\begin{myexample}{Prompt Example for BACE Task}{}
\label{example1}
\ttfamily
    \textbf{[Role]}\\
    You are a top AI assistant specializing in molecular chemistry and drug discovery, proficient in molecular property prediction.\\
    
    \textbf{[Task]}\\
    BACE1 is an aspartic-acid protease important in the pathogenesis of Alzheimer's disease, and in the formation of myelin sheaths...\\
    output "True" or "False".\\
    
    \textbf{[Formatting]}\\
    Place the thought process within {\ttfamily\textcolor{blue}{\textless{}think\textgreater{}\textless{}/think\textgreater{}}} and then conclude your answer with {\ttfamily\textcolor{red}{\textless{}answer\textgreater{}}True/False\textcolor{red}{\textless{}/answer\textgreater{}}}.\\
    
    \textbf{[Example]}\\
    {\ttfamily
    \textcolor{blue}{\textless{}think\textgreater{}}xxxx\textcolor{blue}{\textless{}/think\textgreater{}}\\
    \textcolor{red}{\textless{}answer\textgreater{}}True/False\textcolor{red}{\textless{}/answer\textgreater{}}
    }\\
    
    \textbf{[Few-shot]}\\
    \begin{tabular}{|p{10cm}|c|}
        \hline
        {\scriptsize \ttfamily ClC1=CC(=CC(Cl)=C1NC(=O)C)CNC(=[NH2+1])NC(=O)CN2C3=CC(OC)=CC=C3C=C2} & False \\
        \hline
        {\scriptsize \ttfamily ClC1=CC(=CC(Cl)=C1NC(=O)C)CNC(=[NH2+1])NC(=O)CN2C3=CC(CC)=CC=C3C=C2} & True \\
        \hline
        {\scriptsize \ttfamily ClC1=CC(=CC(Cl)=C1NC(=O)C)CNC(=[NH2+1])NC(=O)CN2C3=CC(F)=CC3C=C2} & False \\
        \hline
    \end{tabular}\\
    \\
    \\
    \textbf{[Molecule]}\\
    {\small \ttfamily ClC1=CC(=CC(Cl)=C1NC(=O)C)CNC(=[NH2+1])NC(=O)CN2C3=C(C=CC=C3)C=C2}\\
    \\
    \includegraphics[width=0.5\textwidth]{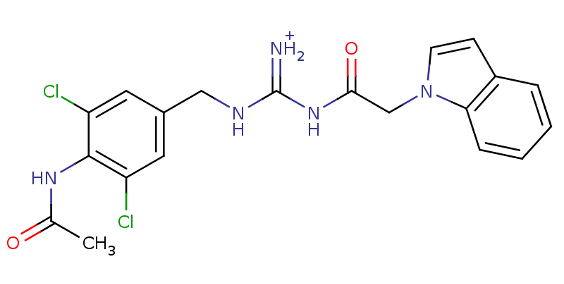}
\end{myexample}

\vspace{1cm}

\begin{myexample}{Prompt for ChemCoT-Based One-Shot Generattion}{}
\ttfamily
    {\ttfamily\textbf{Example Prompt:}}\\
    
    \textless{} PORMPT retrieved from {\ttfamily \underline{OpenMol/ChemCoTDataset}} \textgreater{}\\

    {\ttfamily\textbf{Example Response:}}\\

    \textless{} RESPONSE retrieved from {\ttfamily \underline{OpenMol/ChemCoTDataset}} \textgreater{}\\

    {\ttfamily\textbf{Prompt:}}\\
    
    \textless{} PORMPT likes {\ttfamily Example~1 \textgreater{}}\\    

    {\ttfamily\textbf{Response:}}
\end{myexample}

\begin{myexample}{Prompt for Expert-Guided Task-Specific Generation}{}
\ttfamily
    \textless{} PORMPT likes {\ttfamily Example~1 \textgreater{}}\\    
    
    \textbf{[Expert]}\\

    \textless{} EXPERT KNOWLEDGE refined by {\ttfamily GPT4o \textgreater{}}\\    
\end{myexample}

\vspace{1cm}

\begin{myexample}{Prompt for Logical Soundness Scoring}{}
\ttfamily
You are a professional reasoning-evaluation expert. Your task is to assess the \textbf{logical soundness} of a large language model's chain-of-thought when answering a question, and assign an integer score from \textbf{0 to 10}. Focus strictly on the logical connections between reasoning steps, not on whether the final answer is correct.\\

{\ttfamily\textbf{Input:}}\\
- [Question]: The original question.\\
- [Model Response]: The model's full response, including its chain of thought.\\

{\ttfamily\textbf{Scoring Dimension (Logical Soundness):}}\\
- Do reasoning steps build progressively and refer back to earlier points?\\
- Is each step a reasonable extension of the previous inference?\\
- Is the language coherent, with no contradictions or confusing wording? \\

{\ttfamily\textbf{Scoring Scale (0-10):}}\\
- \textbf{10}: Perfect logical structure; steps are crystal-clear and fully justified.\\
- \textbf{8-9}: Overall logic sound; only minor or negligible leaps/wording issues.\\
- \textbf{6-7}: Main logic correct, but some jumps, insufficient explanation, or minor conflicts.\\
- \textbf{4-5}: Noticeable breaks or missing key inferences, yet some coherent logic remains.\\
- \textbf{2-3}: Most steps lack causality or contradict each other; only sporadic correct parts.\\
- \textbf{0-1}: Virtually no discernible valid reasoning structure.\\


\textbf{Your Task:}\\
Adhering strictly to the rubric above, you must output only a single integer score from 0 to 10. Do not provide any additional explanations, text, or justifications.\\

\textbf{Question:}\\
\textless{} QUESTION \textgreater{}\\

\textbf{Model Response:}\\
\textless{} RESPONSE \textgreater{}\\

\textbf{Output Format:} [integer score]
 
\end{myexample}

\begin{myexample}{Prompt for Accuracy \& Insight Scoring}{}
\ttfamily
You are a professional reasoning-evaluation expert. Your task is to assess the \textbf{accuracy and insight value} of a large language model's chain-of-thought when answering a question, and assign an integer score from \textbf{0 to 10}...\\

{\ttfamily\textbf{Scoring Dimension (Accuracy \& Insight):}}\\
- Are the concepts, formulas, and facts used accurate and appropriate? \\
- Do the reasoning perspective, decomposition approach, or intermediate conclusions provide substantive support or fresh insights for domain experts? \\

{\ttfamily\textbf{Scoring Scale (0-10):}}\\
- \textbf{10}: All methods and facts are completely correct, offering deep and original insights.\\
- \textbf{8-9}: Core content is correct, with only minor detail errors or slightly shallower insights.\\
- \textbf{6-7}: Mostly correct, but with notable secondary errors or average insight depth.\\
- \textbf{4-5}: Mix of correct and incorrect information; limited insight value.\\
- \textbf{2-3}: Most methods/facts are wrong or misused, providing almost no insight.\\
- \textbf{0-1}: Completely incorrect or irrelevant.\\
...
\\
\end{myexample}

\vspace{1cm}

\begin{myexample}{Prompt for Accuracy \& Insight Scoring}{}
\ttfamily
You are a professional reasoning-evaluation expert. Your task is to assess the \textbf{conciseness} of a large language model's chain-of-thought when answering a question, and assign an integer score from \textbf{0 to 10}...\\

\textbf{Scoring Dimension (Conciseness):}\\
- Does the response go straight to the point, avoiding irrelevant or repetitive explanations? \\
- Does it convey the full reasoning with the minimum necessary steps? \\

{\ttfamily\textbf{Scoring Scale (0-10):}}\\
- \textbf{10}: Extremely concise, with no redundant or repetitive statements.\\
- \textbf{8-9}: Generally concise, with only a tiny amount of removable content.\\
- \textbf{6-7}: Noticeable redundant paragraphs or repeated explanations.\\
- \textbf{4-5}: Long-winded and repetitive; key information diluted by noise.\\
- \textbf{2-3}: Large portions are irrelevant or repetitive; core points hard to discern.\\
- \textbf{0-1}: Almost entirely made up of redundant content.\\
...
\\
\end{myexample}

\begin{table}[htbp]
\centering
\caption{Reasoning quality scores across three evaluation dimensions. All scores are on a 0-10 scale.}
\label{tab:reasoning_scores}
\resizebox{\textwidth}{!}{%
\begin{tabular}{lcccc}
\toprule
\textbf{Model} & \textbf{Logical Soundness} & \textbf{Accuracy \& Insight} & \textbf{Conciseness} & \textbf{Average} \\
\midrule
o3-mini & 7.182 & 5.470 & 6.053 & 6.235 \\
DeepSeek-V3.1-Think & 7.395 & 6.517 & 6.257 & 6.723 \\
GPT-4o & 6.698 & 5.916 & 5.653 & 6.089 \\
Qwen2.5-VL-72B-Instruct & 7.641 & 6.241 & 5.492 & 6.458 \\
Qwen2.5-VL-7B-Instruct & 4.517 & 3.259 & 5.079 & 4.285 \\
\rowcolor{gray!15}
\textbf{MPPReasoner (Ours)} & \textbf{8.556} & \textbf{7.039} & \textbf{7.352} & \textbf{7.730} \\
\bottomrule
\end{tabular}}
\end{table}

\section{Reasoning Quality Scores}
\label{appb:quality}
Table~\ref{tab:reasoning_scores} presents the detailed reasoning quality scores across three evaluation dimensions~\citep{sophiavlr1} for all models in our study.
The detailed dimensional analysis reveals several important insights into model capabilities and reasoning patterns~\citep{surveyreasoning,evaluationlanguagemodels,rewardanything,rmr1}. MPPReasoner achieves the highest scores across all three evaluation dimensions, demonstrating comprehensive reasoning excellence. In logical soundness, MPPReasoner scores 8.556, significantly outperforming the best baseline DeepSeek-V3.1-Think at 7.395, indicating superior coherence in step-by-step reasoning flow. For accuracy \& insight, our model achieves 7.039, substantially exceeding DeepSeek-V3.1-Think's 6.517, which demonstrates the effectiveness of chemical principle integration in generating factually correct and insightful analyses.

Examining model category patterns, advanced reasoning models like o3-mini and DeepSeek-V3.1-Think show relatively strong logical soundness but struggle with accuracy \& insight, particularly o3-mini at 5.470, suggesting that general reasoning capabilities cannot substitute for domain-specific knowledge. Large-scale models exhibit mixed performance: Qwen2.5-VL-72B-Instruct achieves decent logical soundness (7.641) but suffers in conciseness (5.492), while the smaller Qwen2.5-VL-7B-Instruct shows consistently poor performance across all dimensions, with particularly low accuracy \& insight at 3.259.
Notably, MPPReasoner maintains balanced excellence across all dimensions, avoiding the trade-offs observed in baseline models. The model's conciseness score of 7.352 is particularly remarkable, as it demonstrates the ability to provide comprehensive chemical reasoning without unnecessary verbosity, a crucial factor for practical applications where chemists need clear and actionable insights.

\section{Implementation Details}
\label{appc:setting}

We implement MPPReasoner based on Qwen2.5-VL-7B-Instruct~\citep{qwen25vl}, configured with a maximum sequence length of 8,192 tokens to accommodate detailed reasoning outputs. Our implementation follows a two-stage training pipeline with carefully tuned hyperparameters for optimal performance.

SFT stage employs 16,000 curated reasoning trajectories over 3 epochs. We use an effective batch size of 16 with a learning rate of 1e-5 and the AdamW optimizer. A linear learning rate scheduler with 3\% warmup ratio ensures stable training convergence.

RL stage utilizes the GRPO algorithm~\citep{grpo} for 500 optimization steps with dynamic sampling~\citep{dapo}t o filter training instances and focus on tractable reasoning examples. We employ a lower learning rate of 1e-6 with weight decay of 1e-2 and KL coefficient of 1e-2 to maintain stability during policy optimization. The rollout configuration generates 5 samples per input with temperature 1.0, using a global batch size of 128 and rollout batch size of 512 for efficient training.The hierarchical reward weights in RLPGR are set as ($\lambda_1$, $\lambda_2$, $\lambda_3$) = (1.0, 0.25, 0.25) for foundation, reasoning, and chemistry layers respectively.

All training is conducted on 8 NVIDIA A100 80GB GPUs with mixed precision~\citep{mixedprecisiontraining} training for memory efficiency. The SFT stage requires approximately 2 hours, while the RL stage takes 12 hours, totaling 14 hours for complete training. During inference, we use temperature 1.0 with top-k sampling (k=5) to generate diverse yet high-quality reasoning paths.

\begin{table}[!htb]
\centering
\caption{Hyperparameters Setting}
\label{tab:setting}
\begin{tabular}{l|c}
\toprule
\textbf{Hyperparameter} & \textbf{Value} \\ 
\midrule
\rowcolor{gray!20}
\multicolumn{2}{c}{\textit{\# Supervised Fine-tuning (SFT)}} \\
GPU Number (A100) & 8 \\
Train Batch Size Per Device & 2 \\
Gradient Accumulation Steps & 4 \\
Learning Rate & 1.0e-5 \\
Number of Train Epochs & 3 \\
LR Scheduler Type & Linear \\
Warmup Ratio & 0.03 \\
\midrule
\rowcolor{gray!20}
\multicolumn{2}{c}{\textit{\# Reinforcement Learning (RL)}} \\
GPU Number (A100) & 8 \\
Learning Rate & 1.0e-6 \\
Weight Decay & 1.0e-2 \\
KL Coefficient & 1.0e-2 \\
Rollout Number & 5 \\
Rollout Temperature & 1.0 \\
Global Batch Size & 128 \\
Rollout Batch Size & 512 \\ 
Micro Batch Size For Update Per Device  & 8 \\
\midrule
\rowcolor{gray!20}
\multicolumn{2}{c}{\textit{\# Inference}} \\
Temperature & 1.0 \\
Top K & 5 \\
Max Tokens & 8,192 \\
\bottomrule
\end{tabular}
\end{table}

\section{More Cases}

\begin{figure}[!htp]
    \centering
    \includegraphics[clip, trim=0cm 0.5cm 0cm 0.5cm, width=0.95\textwidth]{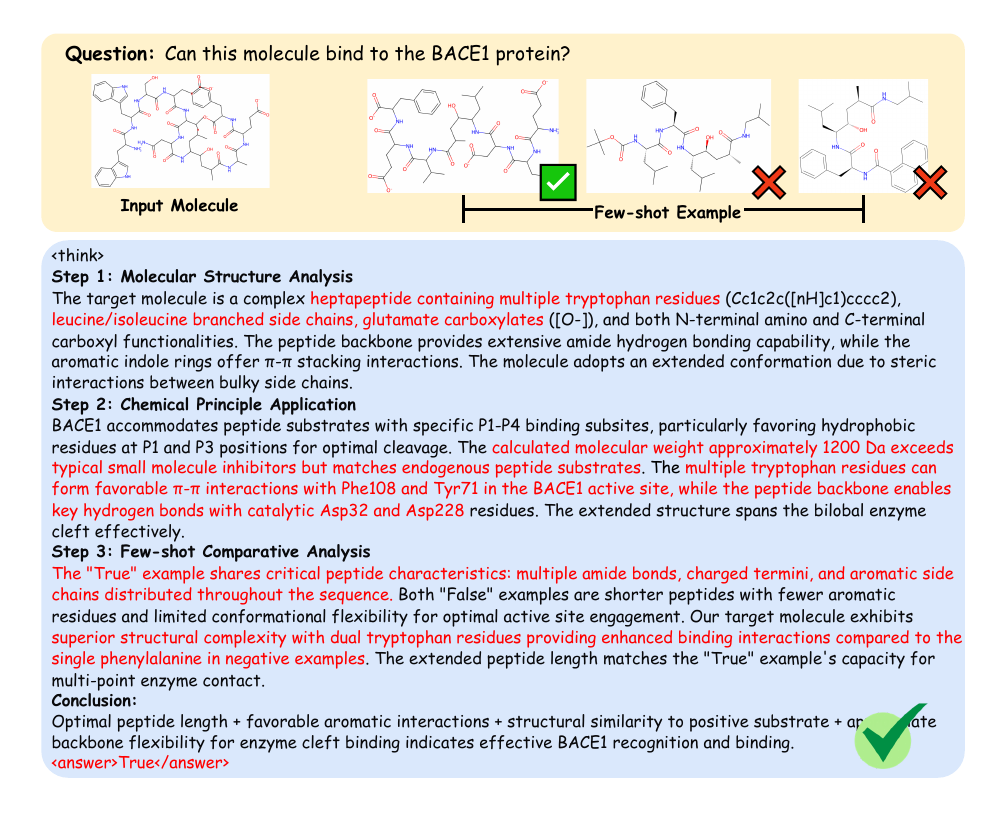}
    \caption{Successful case on MPPReasoner for BACE1 protein binding prediction (ID).}
    \label{fig:more_case1}
\end{figure}

\begin{figure}[!htp]
    \centering
    \includegraphics[clip, trim=0cm 0.5cm 0cm 0.5cm, width=0.95\textwidth]{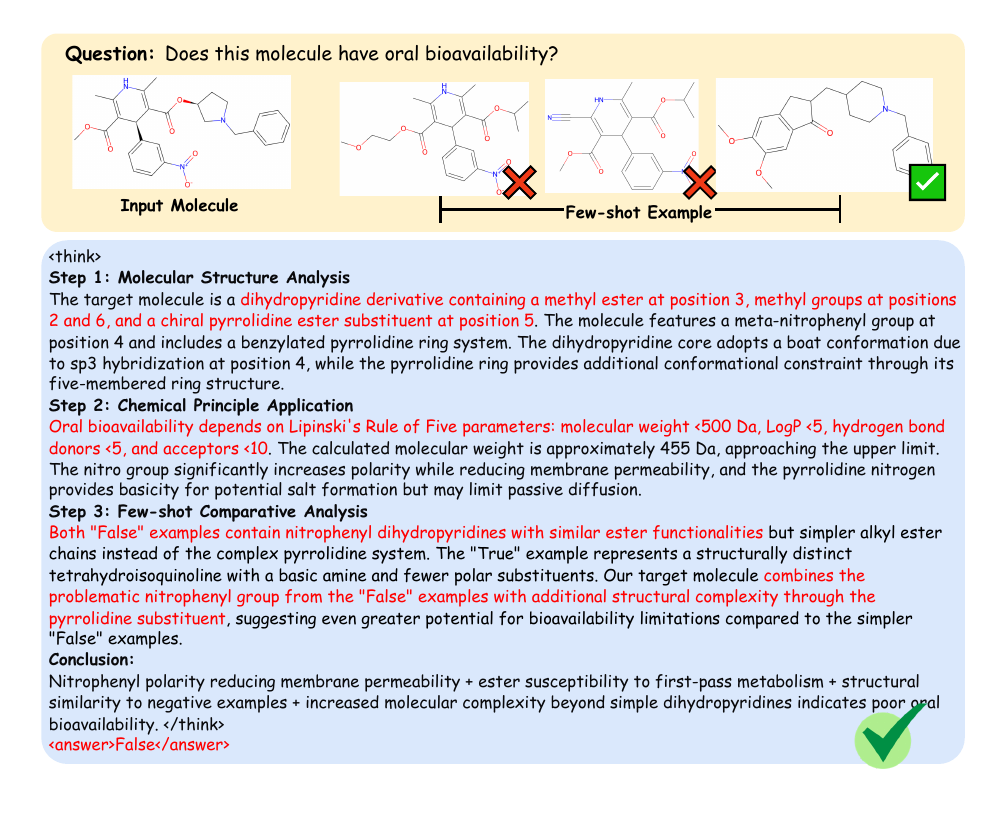}
    \caption{Successful case on MPPReasoner for oral bioavailability prediction (OOD).}
    \label{fig:more_case2}
\end{figure}

\section{Limitations}

While MPPReasoner demonstrates significant advances in chemical reasoning for molecular property prediction, several areas present opportunities for future enhancement:

\begin{itemize}
    \item \textit{Molecular Representation:} Current framework primarily utilizes 1D/2D molecular representations through SMILES and molecular images. Incorporating 3D structural information~\citep{3dllm}, conformational dynamics~\citep{mdsimulations}, and stereochemical effects~\citep{stereochemical} could further enhance prediction accuracy for properties sensitive to spatial arrangements and molecular flexibility.
    
    \item \textit{Computational Efficiency:} The generation of detailed reasoning paths introduces additional computational overhead compared to direct prediction models. This trade-off between interpretability and efficiency may limit scalability for certain high-throughput screening applications~\citep{highscreening}, though the enhanced explainability proves valuable for research and development workflows.
    
    \item \textit{Domain Scope:} The current evaluation focuses on molecular property prediction tasks. Expanding the framework to broader chemical domains such as reaction mechanism prediction~\citep{reactgpt,usptollm}, synthesis planning~\citep{chemdual,designdecision}, and molecular optimization~\citep{moleculeopt,mollm} could demonstrate wider applicability of the chemical reasoning approach.
\end{itemize}

Future work will address these limitations through more efficient architectures, enhanced molecular representations, and broader domain applications while maintaining the interpretability advantages that distinguish our approach.

\end{document}